\documentclass[runningheads]{llncs}
\usepackage{graphicx}
\usepackage{amsmath}
\usepackage{multirow}
\usepackage{makecell}

\usepackage{amsfonts}
\usepackage{amssymb}
\usepackage{subfig}

\usepackage{times}
\usepackage{soul}
\usepackage{url}
\usepackage{algorithm}
\usepackage{algorithmic}
\begin{document}
\title{ApproBiVT: Lead ASR Models to Generalize Better Using Approximated Bias-Variance Tradeoff Guided Early Stopping and Checkpoint Averaging\thanks{Supported by the National Innovation 2030 Major
S\&T Project of China under Grant 2020AAA0104202.}}
%
%
\author{Fangyuan Wang\inst{1}\orcidID{0000-0002-6482-4522} \and Ming Hao\inst{2} \and Yuhai Shi\inst{2} \and
Bo Xu\inst{1,3,4}}
%
%
\institute{Institute of Automation, Chinese Academy of Sciences, Beijing, China \and
Academy of Broadcasting Science, National Radio and Television Administration, Beijing, China \and
School of Future Technology, University of Chinese Academy of Sciences
, Beijing, China \and
School of Artificial Intelligence, University of Chinese Academy of Sciences, Beijing, China\\
\email{\{fangyuan.wang,xubo\}@ia.ac.cn}, 
\email{\{haoming,shiyuhai\}@abs.ac.cn}
}
\maketitle              
\begin{abstract}
The conventional recipe for Automatic Speech Recognition (ASR) models is to 1) train multiple checkpoints on a training set while relying on a validation set to prevent overfitting using early stopping and 2) average several last checkpoints or that of the lowest validation losses to obtain the final model. In this paper, we rethink and update the early stopping and checkpoint averaging from the perspective of the bias-variance tradeoff.
Theoretically, the \emph{bias} and \emph{variance} represent the fitness and variability of a model and the tradeoff of them determines the overall generalization error.
But, it's impractical to evaluate them precisely.
As an alternative, we take the training loss and validation loss as proxies of bias and variance and guide the early stopping and checkpoint averaging using their tradeoff, namely an Approximated Bias-Variance Tradeoff (ApproBiVT). When evaluating with advanced ASR models, our recipe provides 2.5\%-3.7\% 
and 3.1\%-4.6\% CER reduction on the AISHELL-1 and AISHELL-2, respectively
\footnote{The code and sampled unaugmented training sets used in this paper will be public available on GitHub.}.

\end{abstract}

\section{Introduction}

In the past ten years, neural network based End-to-End (E2E) ASR systems have achieved great progress and overwhelming success.
Many advanced network architectures have been proposed and applied in ASR tasks, such as Convolution Neural Networks (CNN)~\cite{Li:Jasper,Kriman2020QuartznetDA,Han2021MultistreamCF}, Recurrent Neural Networks (RNN)~\cite{Chan2016ListenAA,Rao2017ExploringAD}, Self-Attention Networks (SAN) ~\cite{Vaswani2017AttentionIA,Dong2018SpeechTransformerAN,Moritz2020StreamingAS}, and CNN/SAN hybrid networks~\cite{Gulati2020ConformerCT,Zhang2020UnifiedSA,Wu2021U2UT,An2022CUSIDECS,Ren2022ImprovingMS,Wang2022ShiftedCE,Kim2022SqueezeformerAE}.

Regardless of specific model architecture, most state-of-the-art ASR recipes involve two steps: 1) train multiple checkpoints on a training set while using a held-out validation set to decide when to stop training and 2) yield the final model by averaging several last checkpoints or that have the lowest losses on the held-out validation set.
However, this conventional recipe has two downsides.
\begin{figure}[htb]

\begin{minipage}[b]{1.0\linewidth}
  \centering
  \centerline{\includegraphics[width=6.5cm]{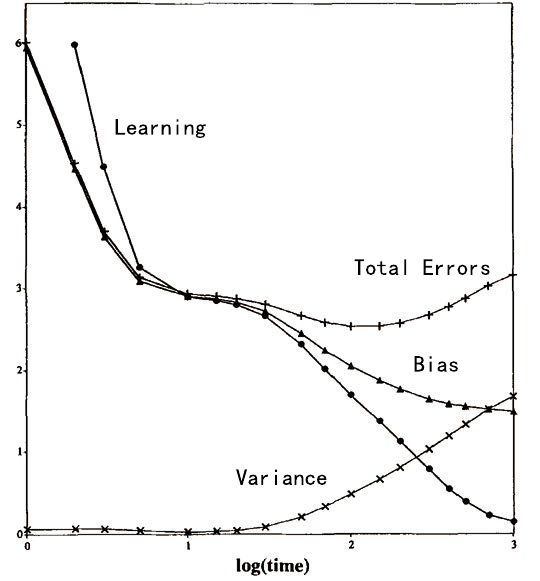}}
\end{minipage}
\caption{ Figure 15 in ~\cite{Geman1992NeuralNA} visualizes the correlation between training loss (Learning), bias, validation loss (Total Errors), and variance. Details about this Figure are in ~\cite{Geman1992NeuralNA}.}
\label{fig:bias-variance}
\end{figure}
For one, the validation loss based early stopping procedure \cite{Morgan1990GeneralizationAP,Reed1993PruningAS,Prechelt1996EarlySW} determines the stop point solely on the validation loss, which is more like a proxy of variance rather than the overall generalization error, without considering the bias explicitly, and may prevent the exploration of more checkpoints of lower bias-variance tradeoff scores.
For another, existing checkpoint averaging procedures either prefer to pick checkpoints of lower biases (last ones in the same training trajectory, the last-\emph{k} (LK) checkpoint averaging scheme )~\cite{Vaswani2017AttentionIA,Popel2018TrainingTF} or that of lower variances (with lower validation losses, the \emph{k}-best validation loss (KBVL) checkpoint averaging scheme)~\cite{Yao2021WeNetPO} to average, discarding the tradeoff of bias and variance, which may limit the power of checkpoint averaging.

\begin{figure*}[!t]
\centering
\subfloat[]{\includegraphics[width=2.42in,height=2.2in]{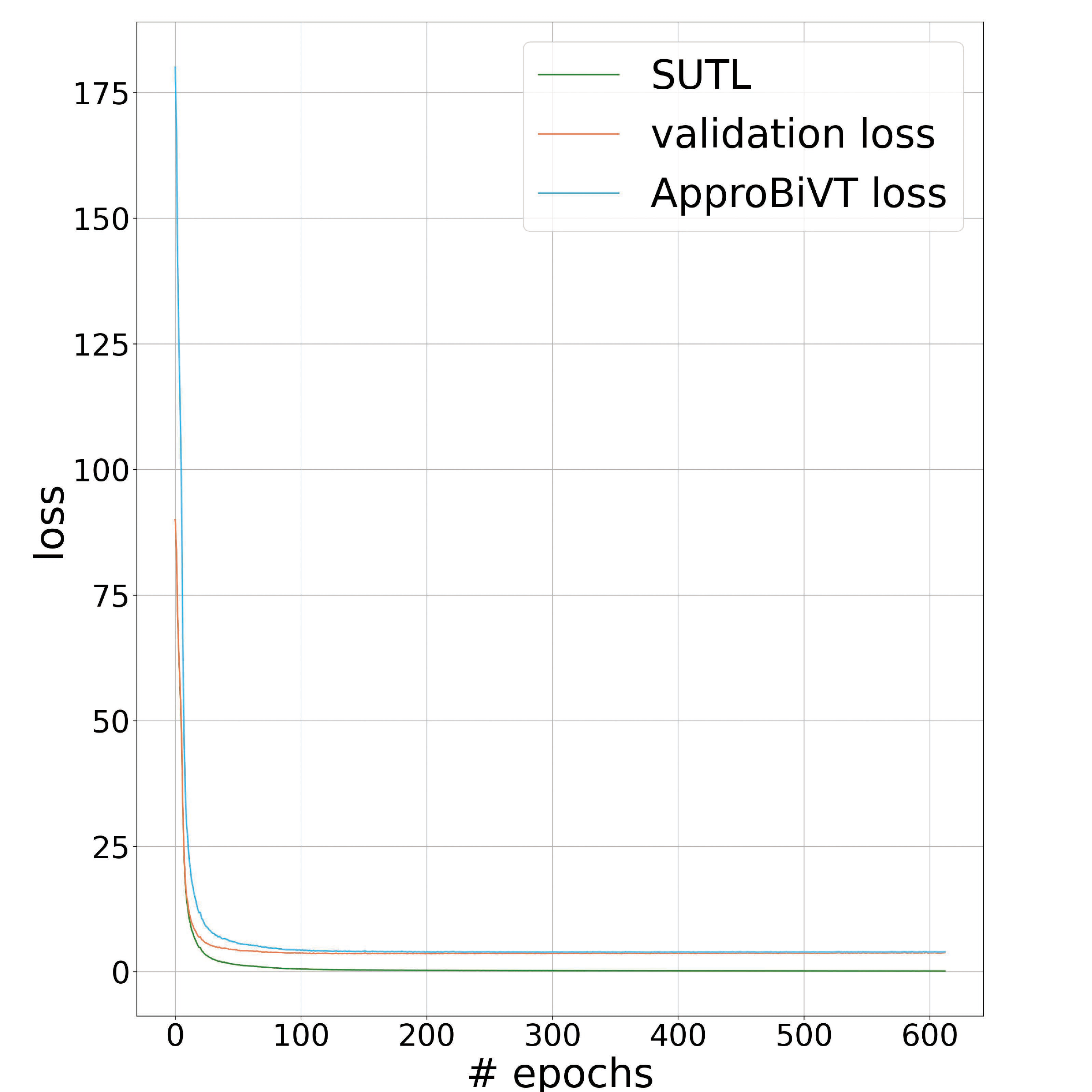}%
\label{fig_first_case}}
\subfloat[]{\includegraphics[width=2.42in,height=2.2in]{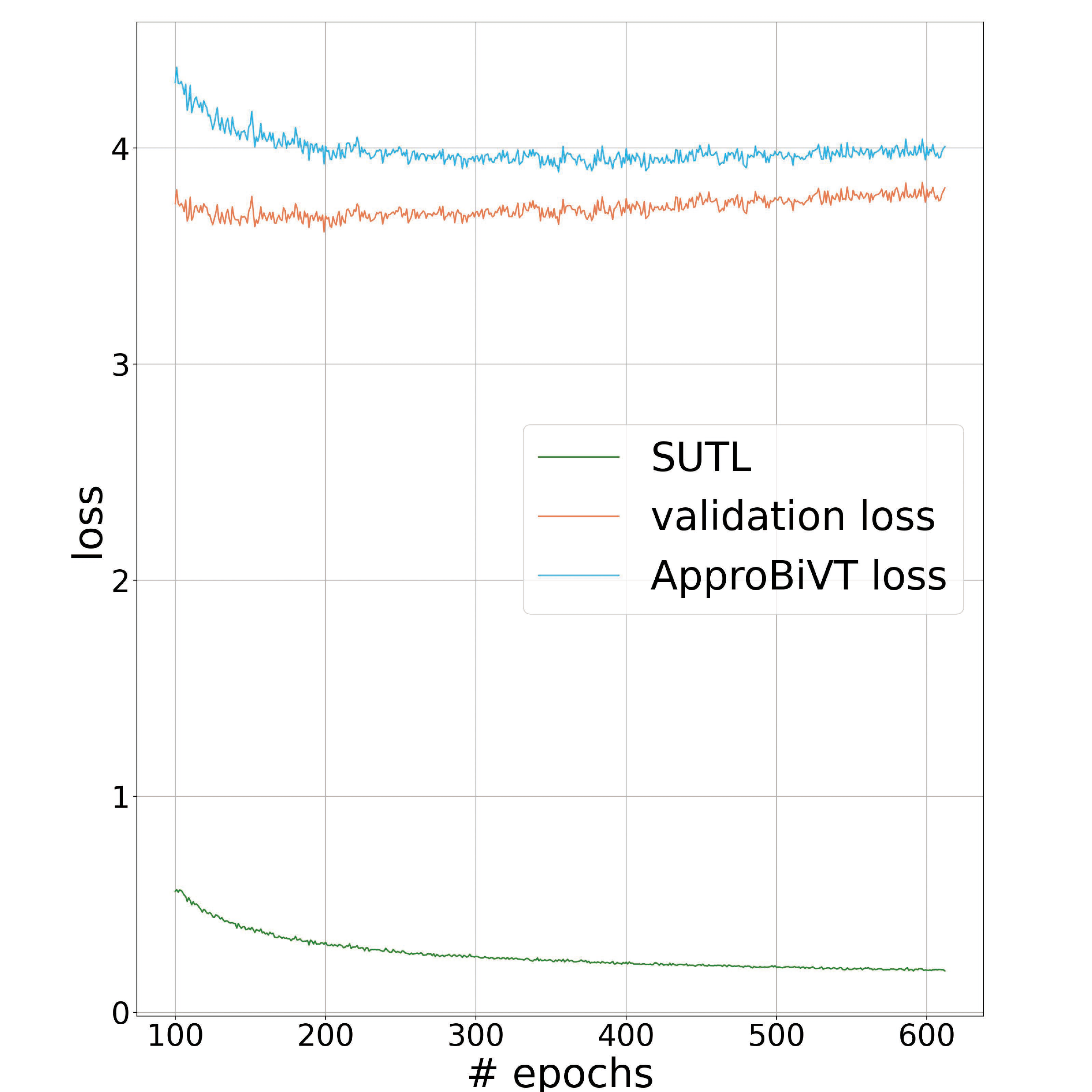}%
\label{fig_second_case}}
\hfil
\subfloat[]{\includegraphics[width=2.42in,height=2.2in]{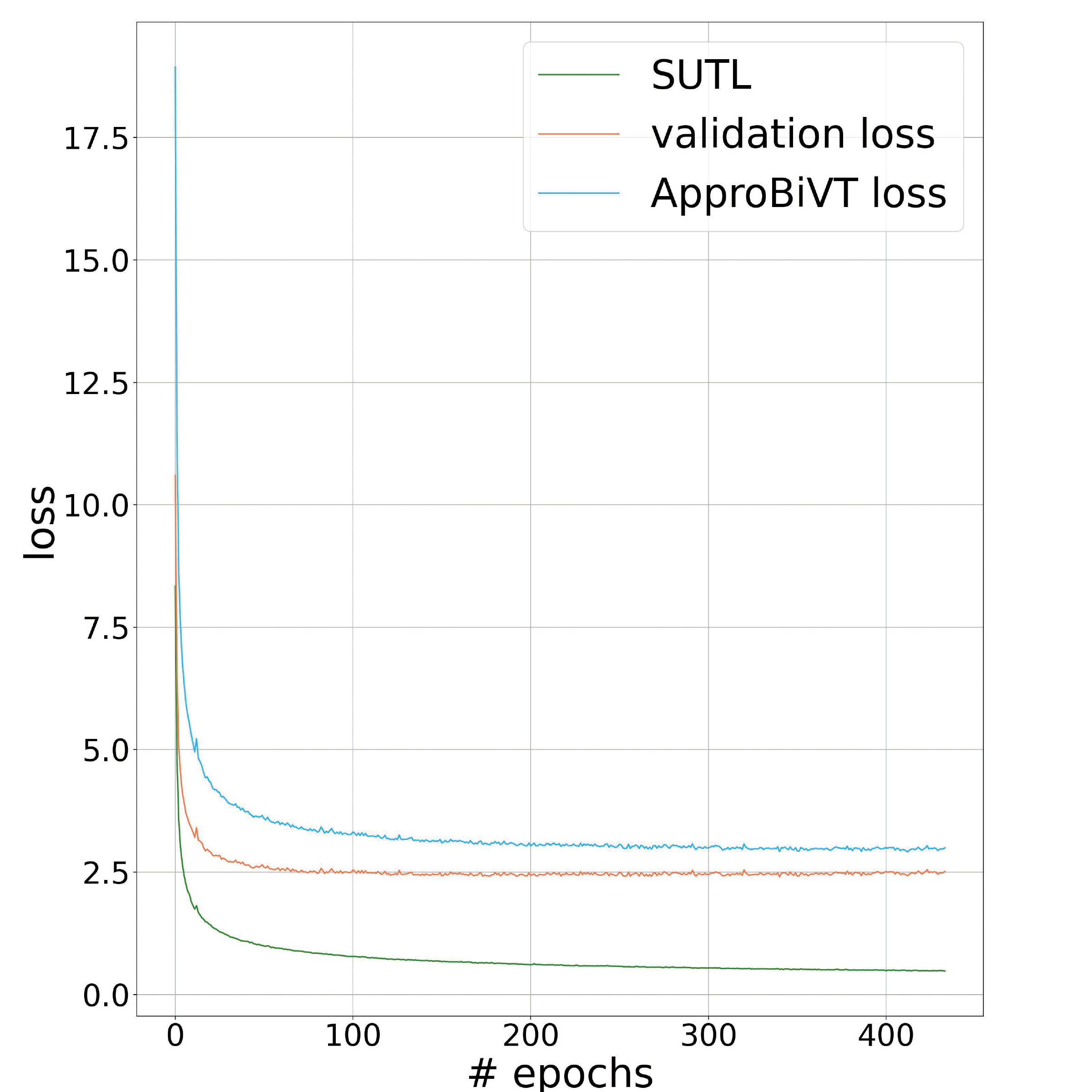}%
\label{fig_third_case}}
\subfloat[]{\includegraphics[width=2.42in,height=2.2in]{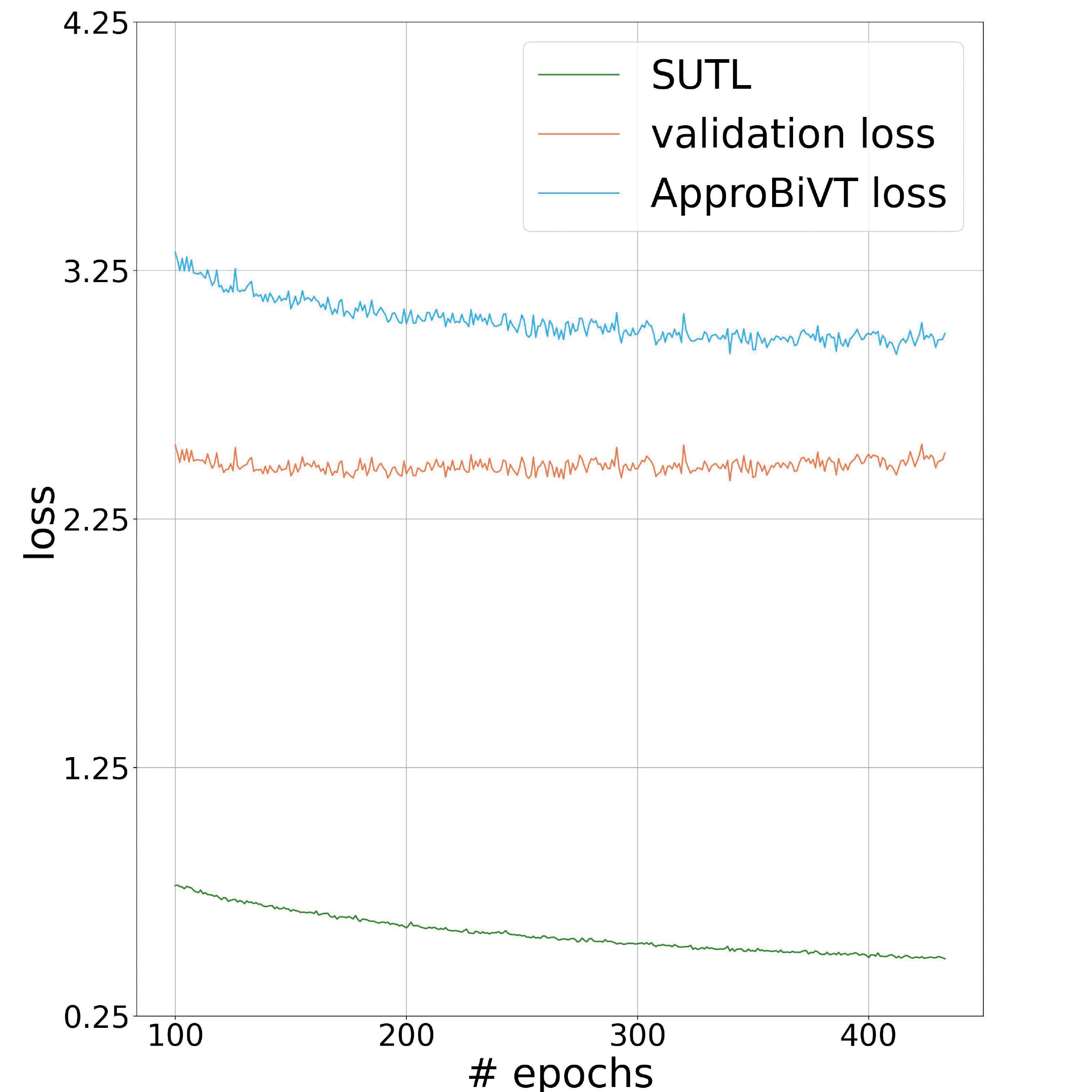}%
\label{fig_fourth_case}}
\caption{The SUTL, validation loss, and ApproBiVT loss curves (overview and zoomed) 
of Conformer on AISHELL-1/-2 (a, b/c, d).
The sampled unaugmented training loss (SUTL, shown in green) illustrates a monotonically decreasing trend.
Whereas the validation loss (shown in red) and ApproBiVT loss (shown in blue) show a trend of falling first and then rising, the difference is that the ApproBiVT loss curve still decreases for many epochs after the validation loss has reached the turning point.
}
\label{fig:approbivt}
\end{figure*}

To lead ASR models to generalize better, we find inspiration from the bias-variance tradeoff~\cite{Geman1992NeuralNA} and propose a reasonable and practical alternative to update the conventional recipe.
1) We argue that it's reasonable to consider the tradeoff of bias and variance.
According to the definition of bias-variance decomposition~\cite{Geman1992NeuralNA},
the generalization error consists of noise, bias, and variance.
The \emph{noise} is the response to the label error in the training set, the \emph{bias} measures how closely the average evaluator matches the target, and the \emph{variance} indicates how much the guess fluctuates for evaluators.
Since the overall generalization error determines the final performance, any recipe only considering bias or variance cannot ensure the best generalization.
Unfortunately, it’s nontrivial to evaluate the bias and variance precisely using the Monte Carlo procedure~\cite{Geman1992NeuralNA}.
2) We propose a reasonable and practical method to approximate the bias-variance tradeoff.
Fig.~\ref{fig:bias-variance} depicts the learning curves of training loss, bias, validation loss, and variance for a neural network. 
We observe that the training loss and bias have a positive correlation, while the trend of validation loss and variance at the end also tends to be positively correlated.
This phenomenon encourages us to take the training loss and validation loss as the proxies of bias and variance and use their summary to approximate the tradeoff of bias and variance.
In particular, we use a Sampled Unaugmented Training Loss (SUTL), instead of the naive training loss, as the proxy of bias to avoid using the total samples in the training set and the distortion of training loss introduced by the data augmentation~\cite{Park2019SpecAugmentAS} and regularization~\cite{Konda2015DropoutAD}.
The core concepts are shown in Fig.~\ref{fig:approbivt} and Fig.~\ref{fig:loss}.

We update the ASR recipe using the Approximated Bias-Variance Tradeoff from two aspects:
1) we use the ApproBiVT score to tell when the training process needs to stop, and our recipe tends to allow training longer until converged according to the ApproBiVT score; 2) we implement a \emph{k}-best ApproBiVT (KBABVT) checkpoint averaging scheme to yield the final model.
Experiments with various ASR models show that our recipe can outperform the conventional ones by 2.5\%-3.7\% on AISHELL-1~\cite{Bu2017AISHELL1AO} and 3.1\%-4.6\% on AISHELL-2~\cite{Du2018AISHELL2TM}.
To our knowledge, this is the first work that uses the training loss and validation loss together to guide the early stopping and checkpoint averaging in ASR recipes.

\section{ApproBiVT Recipe}
\subsection{Bias-Variance Decomposition}
The bias-variance decomposition of mean squared error is
well understood. However, the prominent loss function used in E2E ASR is the cross-entropy. We present the decomposition
for cross-entropy as in~\cite{Heskes1998BiasVarianceDF}. We denote the training set as \emph{D} and the output distribution on a sample \emph{x} of the network trained as $\hat{y}=f(x,D)$. Then let the average output of $\hat{y}$ be $\overline{y}$, that is, 
\begin{equation}
    \overline{y} = \frac{1}{Z}\exp(\mathbb{E}_D[\log\hat{y}]) \\
    \label{eq1}
\end{equation}
where \emph{Z} is a  normalization constant. According to~\cite{Heskes1998BiasVarianceDF}, we have the 
following decomposition for the expected error on the sample $x$ and $y = t(x)$ is the ground truth label:
\begin{equation}
	\begin{aligned}	 
    &\text{\emph{error}} =\mathbb{E}_{x,D}[-y\log\hat{y}]\\
          &=\mathbb{E}_{x,D}\left[-y\log{y}+y\log\frac{y}{\overline{y}}+y\log\frac{\overline{y}}{\hat{y}}\right]\\
          &=\mathbb{E}_{x}[-y\log{y}]+\mathbb{E}_{x}\left[y\log\frac{y}{\overline{y}}\right] 
          + \mathbb{E}_{D}\left[\mathbb{E}_{x}\left[y\log\frac{\overline{y}}{\hat{y}}\right]\right] \\
          &=\mathbb{E}_{x}[-y\log{y}]+ D_{KL}(y,\overline{y})+\mathbb{E}_{D}\left[D_{KL}(\overline{y},\hat{y})\right]\\
          &= \text{\emph{intrinsic noise}} + \text{\emph{bias}} + \text{\emph{variance}}
    \end{aligned}\label{eq2}
\end{equation}
where, the \emph{bias} is the divergence between $\overline{y}$ and $y$ to measure how closely the average evaluator matches the target;
the \emph{variance} is the expected divergence between $\overline{y}$ and each guess $\hat{y}$ to measure how much the guess fluctuates for evaluators. 

\subsection{Approximated Bias-Variance Tradeoff} 
In general, it's nontrivial to evaluate the bias and variance precisely using the Monte Carlo procedure~\cite{Geman1992NeuralNA}. We need large training samples and split them into many independent sets, which is impracticable for most ASR tasks.

According to~\cite{Geman1992NeuralNA}, see Fig.~\ref{fig:bias-variance}, each training iteration will decrease the training loss and bias but increase variance, and the validation loss starts to increase dramatically when the variance grows sharply.
This encourages us to regard the validation loss as the proxy of variance rather than the overall generalization error.
And also, there are some empirical evidences that networks generalize better if they are training longer~\cite{Prechelt1996EarlySW,Hoffer2017TrainLG}.
Since the training loss and bias decrease simultaneously in the training process, we suggest taking the training loss as a proxy of bias.
Borrow the tradeoff format of Eq.~\ref{eq2}, we take the sum of training loss and validation loss as the proxy of the generalization error.

\begin{figure}[H]
\centering
\subfloat[]{\includegraphics[width=2.4in,height=2.2in]{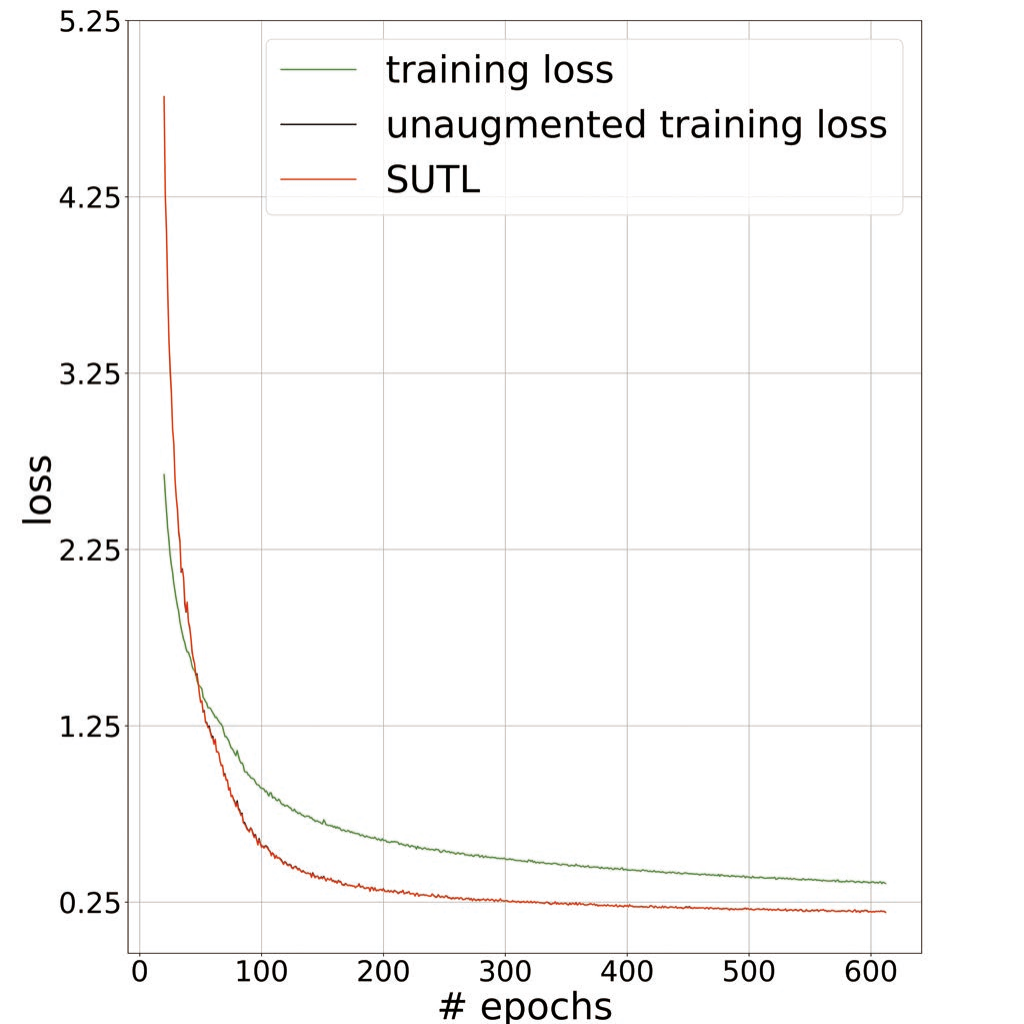}%
\label{fig_first_case}}
\hfil
\subfloat[]{\includegraphics[width=2.4in,height=2.2in]{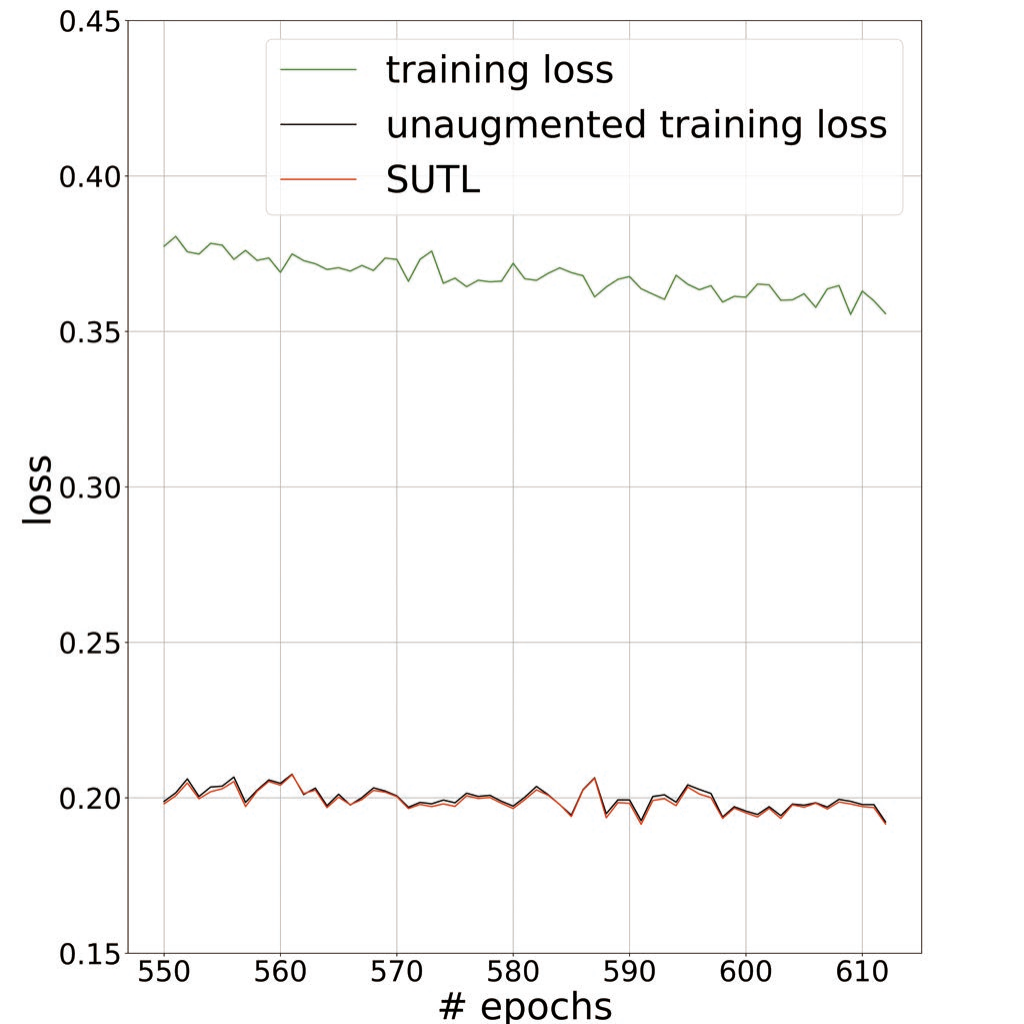}%
\label{fig_third_case}}
\caption{The training loss, SUTL, and unaugmented training loss curves of Conformer on AISHELL-1, (a) overview (depicts from the 20th epoch), (b) zoomed. The divergence between the trend of unaugmented training loss (shown in black) and training loss (shown in green) is significant. In contrast, the trend of sampled unaugmented training loss (SUTL, shown in red) and unaugmented training loss is roughly the same.}
\label{fig:loss}
\end{figure}

However, it may introduce significant ``bias" if we take the naive training loss as the proxy of bias incurred by the difference between the training and validation processes.
Let's revisit the overall training and validation processes.
Typically, the conventional recipe divides the total samples into a training set, a validation set, and a test set.
During training, several augmentation and regularization methods, such as speed perturbation, spectral augmentation~\cite{Park2019SpecAugmentAS}, dropout~\cite{Konda2015DropoutAD} et al., are conducted to prevent overfitting.
Whereas, no additional augmentation or regularization operations are executed in the validation process to ensure the evaluations of validation set and test set are under the same setting. 
To avoid the ``bias" incurred by the augmentation and regularization methods, a natural solution is to take the training set as a validation set and reevaluate on each checkpoint. 
But it's inefficient as the size of a training set is typically much larger than that of a validation set.
In our work, to be precise and efficient, we propose to use a randomly sampled training set that has the same size of validation set to conduct reevaluation on each checkpoint to get the SUTL as the proxy of bias. The loss curves of training loss, unaugmented training loss, and SUTL are illustrated in Fig.~\ref{fig:loss}. Furthermore, we take the sum of SUTL and validation loss as the ApproBiVT:
\begin{equation}
   ApproBiVT \triangleq \text{\emph{SUTL}}+\text{\emph{validation loss}}\\
\label{eq3}
\end{equation}

The overall ApproBiVT recipe is shown in algorithm~\ref{alg:age_paradigm}.

\begin{algorithm}[tb]
    \caption{ApproBiVT recipe}
    \label{alg:age_paradigm}
    \textbf{Preparation}: Construct a sampled unaugmented training set by randomly sampling from the training set.\\
    \textbf{Training}: 
    \begin{algorithmic}[1] 
        \STATE Train the ASR model iteratively, and save a checkpoint once finished an epoch.
        \STATE Evaluate each checkpoint on the sampled unaugmented training set and the validation set to get the SUTL and validation loss, respectively.
        \STATE Calculate the ApproBiVT loss according to Eq.~\ref{eq3}.
        \STATE Conduct early stopping of algorithm~\ref{alg:age_earlystopping_algorithm}.
    \end{algorithmic}
    \textbf{Inference}:    
    \begin{algorithmic}[1] 
        \STATE Conduct \emph{k}-best ApproBiVT checkpoint averaging.
        \STATE Decode using the final model to generate the outputs.
    \end{algorithmic}
\end{algorithm}

\subsection{ApproBiVT-Guided Early Stopping}
We implement a simple ApproBiVT-guided early stopping procedure and stop the training when the ApproBiVT loss monotonically increases for \emph{S} epochs,  see algorithm~\ref{alg:age_earlystopping_algorithm}.
The stop point can be regarded as the inflection point of the ApproBiVT learning curve, which implies that real overfitting may occur.

\subsection{\emph{k}-Best ApproBiVT Checkpoint Averaging}
To further release the power of checkpoint averaging, we introduce a \emph{k}-best ApproBiVT checkpoint averaging procedure to average \emph{k} checkpoints of the lowest ApproBiVT losses in parameter space.
In contrast, the \emph{k}-best validation loss based method prefers to select checkpoints of lower variances.

\section{Experiments}
\label{sec:experiments}
\subsection{Dataset}
We conduct most experiments on AISHELL-1~\cite{Bu2017AISHELL1AO}. It has 150 hours of the training set, 10 hours of the dev set, and 5 hours of the test set.
To test on a large dataset, we verify our recipe on AISHELL-2~\cite{Du2018AISHELL2TM}, which consists of 1000 hours of the training set, 2 hours of the dev set, and 4 hours of the test set.

\subsection{Experimental Setup}
We implement our recipe using the WeNet 2.0 toolkit~\cite{Yao2021WeNetPO} and conduct with 2 NVIDIA GeForce RTX 3090 GPU cards.
The inputs are 80-dim FBANK features with a 25-ms window and a 10-ms shift. 
SpecAug~\cite{Park2019SpecAugmentAS} with two frequency masks with maximum frequency mask (F=10) and two times masks with maximum time mask (T=50) is applied. 
The SpecSub~\cite{Wu2021U2UT} is used in U2++~\cite{Wu2021U2UT} to augment data with Tmax, Tmin, and Nmax set to 30, 0, and 3, respectively.
Two convolution sub-sampling layers with kernel size $3\times3$ and stride 2 are used in the front. 
We use 12 stacked layers of Conformer or an improved variant on the encoder side.
The kernel size of the convolution layer for the Conformer and Blockformer~\cite{Ren2022ImprovingMS} is 15, while 8 for the U2~\cite{Zhang2020UnifiedSA} and U2++~\cite{Wu2021U2UT} with casual convolution.
On the decoder side, a CTC decoder and an Attention decoder of 6 Transformer blocks are used.
The number of attention heads, attention dimension, and feed-forward dimension are set to 4, 256, and 2048, respectively.
We use the Adam optimizer~\cite{Jais2019AdamOA} with the Transformer schedule to train models until they converged.
The value of $S$ is 5 in Algorithm~\ref{alg:age_earlystopping_algorithm}.
For the other hyper-parameters, we just follow their default recipes.

\subsection{Evaluated Models}
We mainly verify our recipe on the Conformer~\cite{Gulati2020ConformerCT} model.
To test whether the gains obtained by the ApproBiVT recipe are additive with other techniques, we also evaluate our recipe with 
U2~\cite{Zhang2020UnifiedSA}, U2++~\cite{Wu2021U2UT}, and Blockformer~\cite{Ren2022ImprovingMS}.

\begin{algorithm}[tb]
    \caption{ApproBiVT-guided early stopping procedure}
    \label{alg:age_earlystopping_algorithm}
    \textbf{Input}: $\{\emph{L}_0,\emph{L}_1,...\}$: the ApproBiVT losses of checkpoints.\\
    \textbf{Parameter}: $\emph{S}$: the maximum number of epochs with monotonically increased ApproBiVT losses.\\
    \textbf{Output}: the stop point $E$
    \begin{algorithmic}[1] 
        \STATE Let $i=S$.
        \WHILE{True}
        \IF {$\emph{L}_i\geq\emph{L}_{i-1},\emph{L}_{i+1}\geq\emph{L}_i,...,\emph{L}_{i-S+1}\geq\emph{L}_{i-S}$} 
        \STATE $\emph{E}$=$i$, break.
        \ELSE 
        \STATE $i+$$+$, continue.
        \ENDIF
        \ENDWHILE
        \STATE \textbf{return} $E$
    \end{algorithmic}
\end{algorithm}

\subsection{Ablation Studies}

\subsubsection{Early Stop But When?}
\begin{figure}[H]
\centering
\subfloat[]{\includegraphics[width=2.4in]{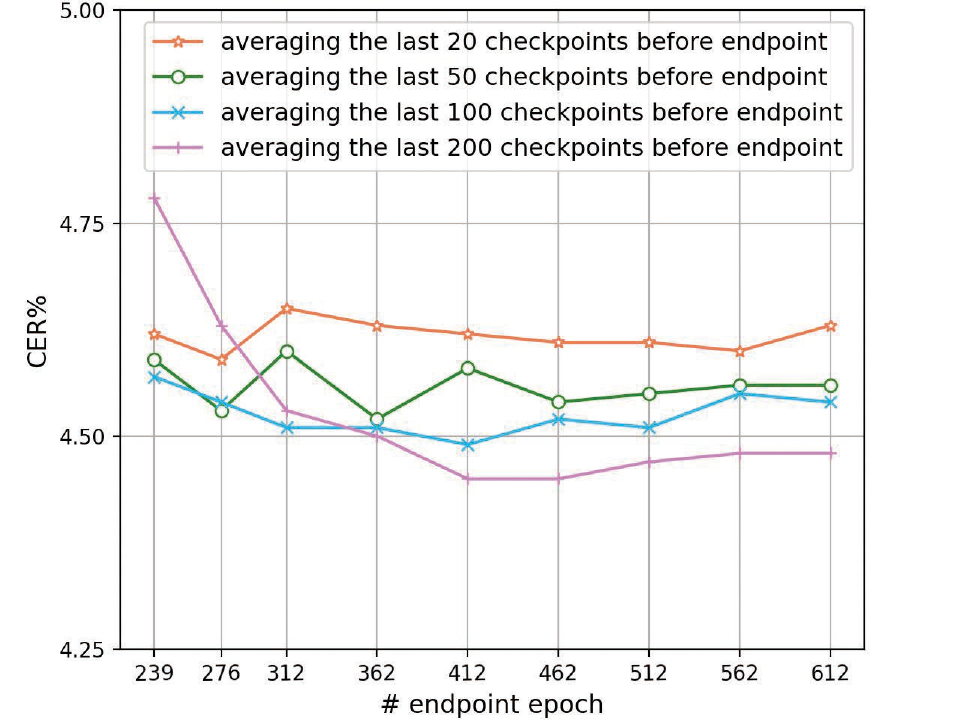}%
\label{fig_first_case}}
\hfil
\subfloat[]{\includegraphics[width=2.4in]{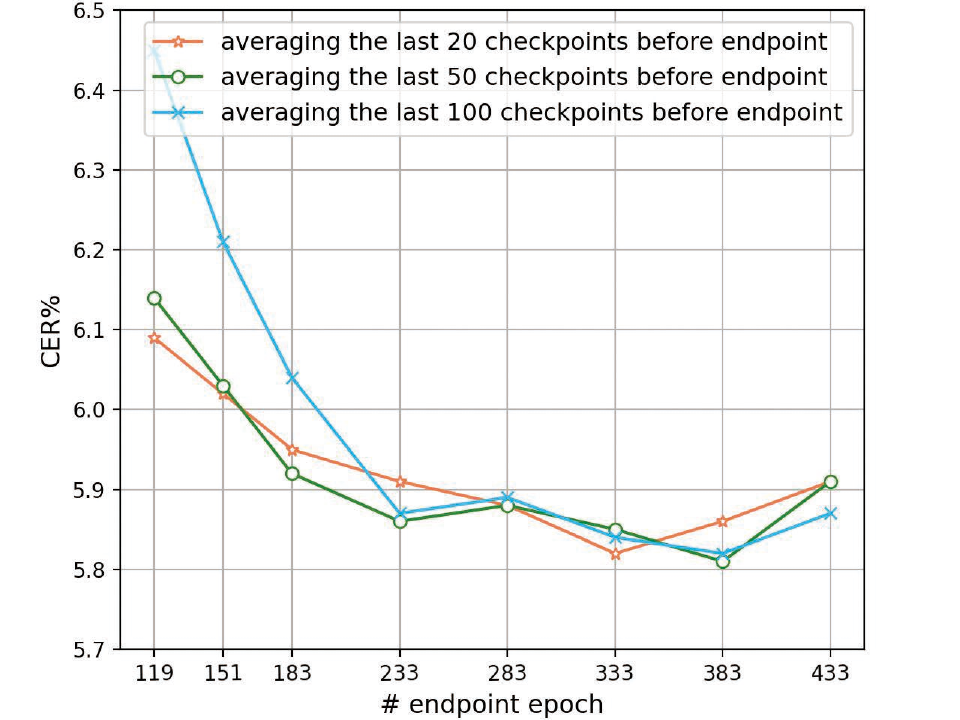}%
\label{fig_second_case}}
\caption{The CERs of Conformer when averaging different numbers of checkpoint at various endpoints using the last-\emph{k} checkpoint averaging procedure, (a) AISHELL-1, (b) AISHELL-2 (IOS test set).}
\label{fig:early_stop}
\end{figure}

First, we explore when stopping training can help to achieve better performance.
We conduct the last-\emph{k} checkpoint averaging using Conformer at various endpoints to detect when the training starts to show
overfitting in the context of checkpoint averaging, see Fig.~\ref{fig:early_stop}.
We can observe that:

1) The overfitting seems to be postponed in the context of checkpoint averaging, especially when averaging more checkpoints. 

2)
Compared to the validation loss based early stopping suggested training 240 and 120 epochs, our recipe allows training more epochs until the ApproBiVT score starts to increase, see Fig.~\ref{fig:approbivt}, thus facilitating to averaging more checkpoints.
And if comparing Fig.~\ref{fig:approbivt} and Fig.~\ref{fig:early_stop}, we can see that the stopping points suggested by our recipe are more reasonable than that of the validation loss based recipe in the context of checkpoint averaging.

\subsubsection{Checkpoint Averaging But How?}
Second, we explore how conducting checkpoint averaging can achieve better performance.
Table~\ref{tab:epochs} lists the results of Conformer with different settings of early stopping and checkpoint averaging procedures.
We can observe that: 

1) With VL guided early stopping, models will converge with 240 epochs, while 613 with ApproBiVT loss guided procedure.

2) With the VL procedure, averaging more checkpoints can not bring additional performance gains.

3) With our early stopping, each averaging procedure can achieve better performance
and averaging more checkpoints seems to hedge out the increased variance introduced by the checkpoints at the last time steps.

4) The best CER, 4.39\%, is achieved by our KBABVT procedure when averaging 100 checkpoints.

5) The KBABVT checkpoint averaging can often achieve better results than the LK and KBVL methods, regardless of how many checkpoints are used to average, which implies the ApproBiVT score is an effective metric for selecting the checkpoints and has verified the necessity of considering bias-variance tradeoff scores.

\begin{table}[t]
\centering
\setlength{\tabcolsep}{4.5mm}
\caption{Comparisons of different early stopping and checkpoint averaging procedures with Conformer on AISHELL-1 test set. The notation of VL is the abbreviation of validation loss. }
\label{tab:epochs}
\begin{tabular}{|c|cccc|}
\hline
 Early   & Avg. & Epoch &Avg. & CER \\
 Stopping  & Procedure &Num.(\#)& Num.(\#) & (\%) \\
\hline
VL &KBVL & $240$ &$20$/$50$& $4.56$/$4.64$  \\
\hline
VL &LK & $240$ &$20$/$50$& $4.62$/$4.64$   \\
\hline
VL &KBABVT & $240$ &$20$/$50$& $4.59$/$4.62$  \\
\hline
ApproBiVT &KBVL & $613$ &\makecell[c]{$20$/$50$\\$100$/$200$}& \makecell[c]{$4.55$/$4.53$\\$4.53$/$4.53$} \\
\hline
ApproBiVT &LK & $613$ &\makecell[c]{$20$/$50$/$100$\\$200$/$300$/$400$}& \makecell[c]{$4.63$/$4.55$/$4.52$\\$4.48$/$4.43$/$4.44$} \\
\hline
ApproBiVT &KBABVT & $613$ &\makecell[c]{$20$/$50$\\$100$/$200$}& \makecell[c]{$4.46$/$4.43$\\\textbf{4.39}/$4.43$} \\
 \hline
\end{tabular}
\end{table}

\begin{table}[t]
\centering
\setlength{\tabcolsep}{3.5mm}
\caption{Evaluation with various models on AISHELL-1.}
\label{tab:aishell-1}
\begin{tabular}{|c|ccccc|}
\hline
\makecell[c]{Model\\Architecture}  &\makecell[c]{Online} &\makecell[c]{Recipe} &\makecell[c]{Epoch\\Num.(\#)} &\makecell[c]{Avg.\\Num.(\#)} & \makecell[c]{CER\\(\%)} \\
\hline
\makecell[c]{U2~\cite{Zhang2020UnifiedSA}}  &Yes   & \makecell[c]{VL+KBVL\\VL+LK\\Ours}  &    
        \makecell[c]{180\\180\\584} & \makecell[c]{20\\20\\100} & \makecell[c]{5.45/5.54\\5.62\\\textbf{5.40}}        \\
\hline
\makecell[c]{U2++~\cite{Wu2021U2UT}}  &Yes   & \makecell[c]{VL+KBVL\\VL+LK\\Ours}  & 
        \makecell[c]{360\\360\\855} & \makecell[c]{30\\30\\200} & \makecell[c]{5.05/5.12\\5.23\\\textbf{4.99}}           \\
\hline
\makecell[c]{Blockformer~\cite{Ren2022ImprovingMS}\footnotemark}  &No   & \makecell[c]{VL+KBVL\\VL+LK\\Ours}  & 
       \makecell[c]{160\\160\\328} & \makecell[c]{40\\40\\100} & \makecell[c]{4.29/4.35\\4.52\\\textbf{4.23}}        \\
 \hline
\end{tabular}
\end{table}
\footnotetext[6]{4.29 are the author reported results. 1 epoch in Blockformer recipe is equivalent to 3 epochs as different data preparation methods used.}

\subsection{AISHELL-1 Results}
Table~\ref{tab:aishell-1} lists the comparisons of several recently proposed state-of-the-art models on AISHELL-1.
In general, if we use the validation loss based early stopping, the checkpoint averaging of KBVL is often superior to the LK scheme.
All these models can obtain additional performance gains using our ApproBiVT recipe, with roughly 2.5\%-3.7\% relative CER reduction compared to their baseline counterparts.
The resulting Blockformer and U2++ models, which attain 4.23\% and 4.99\% CERs, achieve new state-of-the-art for non-streaming and streaming E2E ASR models on AISHELL-1, respectively.
These results confirm that our recipe can be used as an additive technique to enhance ASR models.

\subsection{AISHELL-2 Results}
To test whether the gains of AppriBiVT can scale to a large dataset, we verify our recipe with Conformer on AISHELL-2.
Table~\ref{tab:aishell-2} gives the comparison results between our ApproBiVT recipe and other recipes.
Compared with the baseline recipe of WeNet, our recipe achieves 4.6\%, 3.1\%, and 4.4\% relative CER reduction on the IOS, Android, and Mic test sets, respectively.
This has proved that the ApproBiVT recipe is also an additive technique for large dataset.

\begin{table}[t]
\centering
\setlength{\tabcolsep}{4mm}
\caption{Comparisons of different recipes on AISHELL-2 }
\label{tab:aishell-2}
\begin{tabular}{|c|ccccc|}
\hline
 \makecell[c]{Recipe} &\makecell[c]{Epoch\\Num.(\#)} &\makecell[c]{Avg.\\Num.(\#)} & \makecell[c]{\\IOS} & \makecell[c]{CER(\%)\\Android} & \makecell[c]{\\Mic.} \\
 \hline
        VL+KBVL (WeNet)\footnotemark  & 120 & 20 & 6.09 & 6.69 & 6.09\\
        VL+LK & 120 & 20 & 6.11 &6.62 & 6.12
        \\
        VL+KBABVT & 120 & 20 & 6.09 &6.78 & 6.09
        \\      
        ApproBiVT+KBVL & 434 & 100 & 5.98 &6.52 & 5.98
        \\
        ApproBiVT+LK & 434 & 100 & 5.87 &6.52 & 5.86
        \\
        Ours & 434 & 100 & \textbf{5.81} &\textbf{6.48} & \textbf{5.82}
        \\       
 \hline
\end{tabular}
\end{table}
\footnotetext[7]{The WeNet reported results in https://github.com/wenet-e2e/wenet.}

\subsection{Limitations and Scope}
This work has demonstrated that the ApproBiVT is an additional technique to improve the ASR with no extra inference
cost. It’s no doubt that it typically costs more training time
which is necessary to allow averaging more checkpoints with
low bias and variance tradeoff responses.

\section{Conclusion}
Our results challenge the conventional recipe of early stopping and checkpoint averaging, with no model modification and extra inference computation, we are often able to produce better models using the proposed ApproBiVT recipe.
Our work confirms that training longer can generalize better, gives a reasonable interpretation to indicate how long should be trained and a practical recipe to explore the power of long training.
Our results rewrite the conclusion about checkpoint averaging that no performance improvement can be further squeezed out~\cite{Gao2022RevisitingCA} and provide an effective method to yield better results.
%
%
%
%

\end{document}